\documentclass[10pt]{article}

\usepackage[utf8]{inputenc} 
\usepackage[T1]{fontenc}    

\usepackage[verbose=true,letterpaper]{geometry}
\newgeometry{
    textheight=9in,
    textwidth=6in,
    top=1in,
    headheight=12pt,
    headsep=25pt,
    footskip=30pt
}

\usepackage{lipsum}

\usepackage[utf8]{inputenc}
\usepackage{mathrsfs,amsthm,xcolor,verbatim,bbm,amsmath,amsfonts,amssymb,nicefrac,enumitem}
\usepackage{hyperref,bm,mathtools,xparse,etoolbox}
\usepackage[capitalise,sort]{cleveref} 

\usepackage{textcomp}
\usepackage{sidecap}

\usepackage{booktabs,multirow}
\usepackage[pagewise]{lineno}
\usepackage{algorithm}
\usepackage{algpseudocode}
\usepackage{cancel}
\usepackage{soul,xcolor}
\usepackage{graphicx}
\usepackage{setspace}
\usepackage{pifont}
\usepackage{todonotes}
\usepackage{ulem,marvosym}
\usepackage{caption}
\usepackage{subcaption}
\usepackage{enumitem,kantlipsum}

\algrenewcommand\textproc{}

\newcommand{\bx}{\boldsymbol{x}}
\newcommand{\by}{\boldsymbol{y}}

\newcommand{\bu}{\boldsymbol{u}}
\newcommand{\bv}{\boldsymbol{v}}
\newcommand{\mX}{\mathcal{X}}
\newcommand{\mY}{\mathcal{Y}}

\newcommand{\veps}{\varepsilon}

\newcommand*{\dif}{\mathop{}\!\mathrm{d}}

\setstcolor{red}
\newcommand{\cmark}{\ding{51}}%
\usepackage{ulem}

\DeclareMathOperator*{\argmin}{arg\,min}

\crefname{enumi}{item}{items}

\crefname{figure}{Figure}{Figures}
\crefname{equation}{}{}
\crefname{subsection}{Subsection}{Subsections}

\crefname{case}{Case}{Cases}
\crefname{cor}{Corollary}{Corollaries}
\usepackage{tikz}
\usetikzlibrary{matrix,chains,positioning,decorations.pathreplacing,arrows}
\usetikzlibrary{shapes,arrows}
\tikzset{
	font={\fontsize{9pt}{12}\selectfont}}

\ExplSyntaxOn

\bool_new:N \g_noteobserve

\NewDocumentCommand{\setnote}{}{
  \bool_gset_true:N \g_noteobserve
}

\NewDocumentCommand{\setobserve}{}{
  \bool_gset_false:N \g_noteobserve
}

\NewDocumentCommand{\nobs}{ o }{
  \IfValueT{#1}{
    \str_if_eq:noTF {note} {#1} {
      \bool_gset_true:N \g_noteobserve
    } {
      \str_if_eq:noTF {Note} {#1} {
        \bool_gset_true:N \g_noteobserve
      } {
        \bool_gset_false:N \g_noteobserve
      }
    }
  }
  \bool_if:nTF { \g_noteobserve } {
    \bool_gset_false:N \g_noteobserve
    note
  } {
    \bool_gset_true:N \g_noteobserve
    observe
  }
  \IfValueF{#1}{~}
}

\NewDocumentCommand{\Nobs}{ o }{
  \IfValueT{#1}{
    \str_if_eq:noTF {note} {#1} {
      \bool_gset_true:N \g_noteobserve
    } {
      \str_if_eq:noTF {Note} {#1} {
        \bool_gset_true:N \g_noteobserve
      } {
        \bool_gset_false:N \g_noteobserve
      }
    }
  }
  \bool_if:nTF { \g_noteobserve } {
    \bool_gset_false:N \g_noteobserve
    Note
  } {
    \bool_gset_true:N \g_noteobserve
    Observe
  }
  \IfValueF{#1}{~}
}

\int_new:N \g_furthermore

\NewDocumentCommand{\Moreover}{ o o }{
  \IfValueT{#1}{
    \str_case:nn {#1} {
      {Furthermore} {\int_set:Nn {\g_furthermore} {0}}
      {Moreover} {\int_set:Nn {\g_furthermore} {1}}
      {In~addition} {\int_set:Nn {\g_furthermore} {2}}
      {note} {\bool_gset_true:N \g_noteobserve}
      {observe} {\bool_gset_false:N \g_noteobserve}
    }
    \IfValueT{#2}{
      \str_case:nn {#2} {
        {Furthermore} {\int_set:Nn {\g_furthermore} {0}}
        {Moreover} {\int_set:Nn {\g_furthermore} {1}}
        {In~addition} {\int_set:Nn {\g_furthermore} {2}}
        {note} {\bool_gset_true:N \g_noteobserve}
        {observe} {\bool_gset_false:N \g_noteobserve}
      }
    }
  }
  \int_case:nn { \int_mod:nn {\g_furthermore} {3} } {
    { 0 } { Furthermore,~\nobs that}
    { 1 } { Moreover,~\nobs that}
    { 2 } { In~addition,~\nobs that}
  }
  \int_incr:N \g_furthermore
  \IfValueF{#1}{~}
}

\bool_new:N \g_hencetherefore

\NewDocumentCommand{\hence}{}{
  \bool_if:nTF { \g_hencetherefore } {
    \bool_gset_false:N \g_hencetherefore
    hence~
  } {
    \bool_gset_true:N \g_hencetherefore
    therefore~
  }
}

\NewDocumentCommand{\Hence}{}{
  \bool_if:nTF { \g_hencetherefore } {
    \bool_gset_false:N \g_hencetherefore
    Hence,~we~obtain~
  } {
    \bool_gset_true:N \g_hencetherefore
    Therefore,~we~obtain~
  }
}

\seq_new:N \g_cflist_loaded
\seq_new:N \g_cflist_pending

\NewDocumentCommand{\cfadd}{ m }
{
	\seq_if_in:NnF \g_cflist_loaded { #1 } {
		\seq_if_in:NnF \g_cflist_pending { #1 } {
			\seq_gput_right:Nn \g_cflist_pending { #1 }
		}
	}
}

\NewDocumentCommand{\cfconsiderloaded}{ m }{
	\seq_gput_right:Nn \g_cflist_loaded {#1}
}

\NewDocumentCommand{\cfremove}{ m }
{
	\seq_gremove_all:Nn \g_cflist_pending { #1 }
}

\NewDocumentCommand{\cfload}{ o }
{
	\seq_if_empty:NTF \g_cflist_pending {\unskip} {
		(cf.\ \cref{\seq_use:Nn \g_cflist_pending {,}})\IfValueTF{#1}{#1~}{\unskip}
		\seq_gconcat:NNN \g_cflist_loaded \g_cflist_loaded \g_cflist_pending
		\seq_gclear:N \g_cflist_pending
	}
}

\NewDocumentCommand{\cfclear} {} {
	\seq_gclear:N \g_cflist_loaded
	\seq_gclear:N \g_cflist_pending
}

\NewDocumentCommand{\cfout}{ o }
{
	\seq_if_empty:NTF \g_cflist_pending {\unskip} {
		(cf.\ \cref{\seq_use:Nn \g_cflist_pending {,}})\IfValueTF{#1}{#1~}{\unskip}
		\seq_gclear:N \g_cflist_pending
	}
}

\NewDocumentCommand{\ifnocf} { m } {
	\seq_if_empty:NT \g_cflist_pending { #1 }
}

\ExplSyntaxOff

\NewDocumentEnvironment{cproof}{m}
{\begin{proof}[Proof of \cref{#1}]}%
	{\noindent The proof of \cref{#1} is thus complete.
\end{proof}}

\NewDocumentEnvironment{cproof2}{m}
{\begin{proof}[Proof of \cref{#1}]}%
	{\noindent This completes the proof of \cref{#1}.
\end{proof}}

\theoremstyle{plain} 

\theoremstyle{definition} 

\newtheorem{remark}{Remark}[section]

\title{Fast Gradient Computation for Gromov-Wasserstein Distance}
\author{Wei Zhang$^\dag$, Zihao Wang\thanks{Corresponding to Zihao Wang \texttt{zwanggc@cse.ust.hk}}, Jie Fan$^\dag$, Hao Wu$^\dag$, Yong Zhang$^\ddag$\\
$\dag$Department of Mathematical Sciences, Tsinghua University \\
$*$Department of Computer Science and Engineering, HKUST \\
$\ddag$BNRist, RIIT, Institute of Internet Industry, \\ 
Department of Computer Science and Technology,
Tsinghua University \\}

\begin{document}

\maketitle

\begin{abstract}
The Gromov-Wasserstein distance is a notable extension of optimal transport. In contrast to the classic Wasserstein distance, it solves a quadratic assignment problem that minimizes the pair-wise distance distortion under the transportation of distributions and thus could apply to distributions in different spaces. These properties make Gromov-Wasserstein widely applicable to many fields, such as computer graphics and machine learning. However, the computation of the Gromov-Wasserstein distance and transport plan is expensive. The well-known Entropic Gromov-Wasserstein approach has a cubic complexity since the matrix multiplication operations need to be repeated in computing the gradient of Gromov-Wasserstein loss. This becomes a key bottleneck of the method. Currently, existing methods accelerate the computation focus on sampling and approximation, which leads to low accuracy or incomplete transport plan. In this work, we propose a novel method to accelerate accurate gradient computation by dynamic programming techniques, reducing the complexity from cubic to quadratic. In this way, the original computational bottleneck is broken and the new entropic solution can be obtained with total quadratic time, which is almost optimal complexity. Furthermore, it can be extended to some variants easily. Extensive experiments validate the efficiency and effectiveness of our method.
\end{abstract}

%
%
%

\section{Introduction}

The Gromov-Wasserstein (GW) distance~\cite{memoliGromovWassersteinDistances2011a}, as an important member of Optimal Transport~\cite{santambrogioOptimalTransportApplied2015,peyreComputationalOptimalTransport2019}, is a powerful tool
for distribution comparison. It is related to the Gromov-Hausdorff (GH) distance~\cite{gromovMetricStructuresRiemannian2007}, a fundamental distance in metric geometry that measures how far two metric spaces are from being isometric~\cite{buragoCourseMetricGeometry2001}.
Specifically, it measures the minimal distortion of pair-wise geodesic distances under the transport plan between two probabilistic distributions, even defined on different underlying spaces. Inherited from GH distance, GW is invariant to translation, rotation, and reflection of metric space. In this way, GW has particular advantages to applications that require to preserving geometry structures including computer graphics~\cite{solomonEntropicMetricAlignment2016a,peyreGromovWassersteinAveragingKernel2016,memoliSpectralGromovWassersteinDistances2009}, natural language processing~\cite{alvarez-melisGromovWassersteinAlignmentWord2018a}, graph factorization and clustering~\cite{xuGromovWassersteinFactorizationModels2020,chowdhuryGeneralizedSpectralClustering2021}, and machine learning~\cite{xuLearningAutoencodersRelational2020,bunneLearningGenerativeModels2019a}.
Moreover, variants of GW distance have been proposed for wider applications.
For example,
Unbalanced GW (UGW) extends the comparison from probabilistic distributions to positive measures~\cite{sejourneUnbalancedGromovWasserstein2021}.
Fused GW (FGW) combines the GW and Wasserstein distances by interpolating their objectives, which is shown to be particularly effective for networks~\cite{vayerFusedGromovWassersteinDistance2020a,titouanOptimalTransportStructured2019} and cross-domain distributions~\cite{petricmareticGOTOptimalTransport2019,li2022gromov}.

The computation of the Gromov-Wasserstein distance boils down to solving a non-convex quadratic assignment problem that is NP-hard~\cite{koopmansAssignmentProblemsLocation1957}. For this, some numerical methods built on relaxations have been developed, including convex relaxations~\cite{kezurerTightRelaxationQuadratic2015, solomonSoftMapsSurfaces2012, solomonDirichletEnergyAnalysis2013}, eigenvalue relaxations~\cite{leordeanuSpectralTechniqueCorrespondence2005}, etc. Nevertheless, these methods often require a large number of relaxed variables (for instance, $N^2 \times N^2$ variables in~\cite{kezurerTightRelaxationQuadratic2015} where $N$ is the number of discrete points of two spaces), resulting in high computational complexity. And they frequently provide unsatisfying solutions, especially in the presence of a symmetric metric matrix~\cite{peyreGromovWassersteinAveragingKernel2016}.
Entropic GW is a seminal and now most popular work to compute GW distance from another perspective, which minimizes GW objective with an entropy regularization term~\cite{solomonEntropicMetricAlignment2016a,peyreGromovWassersteinAveragingKernel2016}. In contrast to the non-entropy-based method mentioned above, it exhibits global convergence without removing constraints and offers a more concise computation. Moreover, it can adapt to solve GW variants, such as FGW~\cite{vayerFusedGromovWassersteinDistance2020a} and UGW~\cite{sejourneUnbalancedGromovWasserstein2021}. In each iteration of it, one first computes the GW gradient with matrix multiplications in $O(N^3)$ time, which dominates the total complexity, and then solves the subproblem by the Sinkhorn algorithm~\cite{cuturiSinkhornDistancesLightspeed2013a} in $O(N^2)$ time.

The computational cost of entropic GW remains unsatisfactory in large-scale scenarios. There are various methods to accelerate it (see Table~\ref{tab:method-comparison} for the comparison).
Scalable Gromov-Wasserstein Learning method (S-GWL)~\cite{xuScalableGromovWassersteinLearning2019b} assumes the hierarchical structure of the transportation plan and conducts the entropic GW algorithm in a multi-scale scheme. 
Anchor Energy distance (AE)~\cite{satoFastRobustComparison2020} approximates the quadratic assignment problem by two nested linear assignment problems.  Sampled Gromov-Wasserstein (SaGroW)~\cite{kerdoncuffSampledGromovWasserstein2021} and the importance sparsification method (Spar-GW)~\cite{liEfficientApproximationGromovWasserstein2023} approximate the original problem by sub-sampling original distributions. Low-Rank Gromov-Wasserstein (LR-GW)~\cite{scetbonLineartimeGromovWasserstein2022} assumes the low-rank structures in the distance matrices and transport plan. However, these approximation or sampling methods always lead to low accurate GW distance and incomplete GW transport plans. On the other hand, closed-form solutions for GW can be found on some specific structures, such as 1-dimensional space (Sliced GW)~\cite{titouanSlicedGromovWasserstein2019a} and tree (FlowAlign)~\cite{leFlowbasedAlignmentApproaches2021}. 
\begin{table}[]
\centering
\caption{The comparison of different methods for the computation of GW metric and its variants.}~\label{tab:method-comparison}
\begin{tabular}{lll}
\toprule
Method                                                       & Complexity         & Exact and Full-sized Plan \\ \midrule
\multicolumn{3}{l}{\underline{\textit{Entropic GW and its approximations}}}\\
Entropic GW~\cite{peyreGromovWassersteinAveragingKernel2016} & $O(N^3)$           & \cmark                \\
S-GWL~\cite{xuScalableGromovWassersteinLearning2019b}        & $O(N^2 \log N )$   & not exact             \\
SaGroW~\cite{kerdoncuffSampledGromovWasserstein2021}         & $O(N^2(s+\log N))$ & not full-sized        \\
Spar-GW~\cite{liEfficientApproximationGromovWasserstein2023} & $O(N^2 + s^2)$     & not full-sized        \\
LR-GW~\cite{scetbonLineartimeGromovWasserstein2022}          & $O(Nr^2d^2)$       & not exact         \\
AE~\cite{satoFastRobustComparison2020}                       & $O(N^2 \log N)$    & not exact \\\midrule
\multicolumn{3}{l}{\underline{\textit{GW on special structures}}}\\
Sliced GW~\cite{titouanSlicedGromovWasserstein2019a}          & $O(N^2)$           & 1D space only \\
FlowAlign~\cite{leFlowbasedAlignmentApproaches2021}          & $O(N^2)$           & tree only \\ \midrule
FGC-GW  (This work)                                              & $O(N^2)$           & \cmark                \\\bottomrule
\end{tabular}
\end{table}

In this work, we propose a novel acceleration method to conduct the accurate Entropic GW. The key of the method is to utilize the structure of the distance matrices and dynamic programming techniques to reduce the complexity of the matrix multiplication from $O(N^3)$ time to $O(N^2)$. It directly speeds up the gradient computation that was once the bottleneck in efficiency, which is somehow inspired by the Fast Sinkhorn algorithm~\cite{liaoFastSinkhornAlgorithm2022,liaoFastSinkhornII2024}and fast algorithms for matrix multiplication \cite{indykFasterLinearAlgebra2022a,koevMatricesDisplacementStructure1999}. Therefore, we name the method Fast Gradient Computation for GW metric (FGC-GW). In contrast to the previous methods, FGC-GW presents a rare combination of three advantages: (1) running with low time complexity, (2) computing accurate metrics, (3) inducing exact and full-sized plans. Moreover, it can be extended to some popular GW variants, including FGW and UGW. We call these extensions FGC-FGW and FGC-UGW, respectively.

This paper is organized as follows: Section~\ref{sec:2} briefly reviews the GW distance and the entropic GW algorithm. In Section~\ref{sec:3}. we present the Fast Gradient Computation for GW metric and generalize it to higher dimensions. Extensive experiments in Section~\ref{sec:5} highlight the efficiency, accuracy, and effectiveness of FGC-GW and FGC-FGW. Finally, we conclude the work in Section~\ref{sec:conclusion}.

\section{Gromov-Wasserstein Distance}
\label{sec:2}
Given two probability density functions $u_{\mX}(x)$ and $v_{\mY}(y)$ defined in metric spaces $\mX$ and $\mY$ respectively, the Gromov-Wasserstein distance~\cite{memoliGromovWassersteinDistances2011a} is defined as:
\begin{equation}\label{GW}
    \begin{split}
        &GW^2(u_{\mX}(x),d_{\mX},v_{\mY}(y),d_{\mY}) = \\
        & \qquad \inf_{ \gamma(x,y) \in \mathcal{S}} \int_{\mX^2 \times \mY^2} |d_{\mX}(x,x')-d_{\mY}(y,y')|^2  \gamma(x,y) \gamma(x',y') \dif x \dif x' \dif y \dif y',\\
        &\mathcal{S} = \left\{  \gamma(x,y) \  \Big| \int_{\mY}  \gamma(x,y) \dif y = u_{\mX}(x), \int_{\mX}  \gamma(x,y) \dif x = v_{\mY}(y) \right\}.
    \end{split}
\end{equation}
Here $d_\mX$ and $d_\mY$ are arbitrary metrics on $\mX$ and $\mY$. Usually, the $k$-th power of distances is preferable \cite{peyreComputationalOptimalTransport2019}. For example, in 1D space, 
\begin{equation*}\label{eq:uniform-spacing}
    d_\mX(x,x') =  \lvert x-x' \lvert^{k_{\mX}}, \quad d_\mX(y,y') =\lvert y-y' \lvert^{k_{\mY}},
\end{equation*}
where $k_{\mX},k_{\mY} \in \mathbb{N}^+$.
For numerical computation, we discretize two probability distributions on two uniform grids with spaces of $h_{\mX}$ and $h_{\mY}$. Then, the discrete distributions $\bu_\mX$ and $\bu_\mY$ can be represented by vectors,
\begin{equation*}
    \bu_{\mX} = (u_1,u_2,\cdots,u_M),  \quad  \bv_{\mX} = (v_1,v_2,\cdots,v_N).
\end{equation*} 
In this paper, our discussion is general for any $k_{\mX}$ and $k_{\mY}$. For the sake of simplicity, we assume $k_{\mX}= k_{\mY}=k$. Naturally, there are two symmetric distance matrices
\begin{equation}\label{1d distance matrices}
    \begin{split}
         D_{\mX}&=[d^{\mX}_{ij}]_{M\times M}, \quad  d^{\mX}_{ij} = h_\mX^k \lvert i-j \lvert^k, \\
         D_{\mY}&=[d^{\mY}_{pq}]_{N\times N}, \quad d^{\mY}_{pq} = h_\mY^k \lvert p-q\lvert^k.
    \end{split}
\end{equation}

Therefore, the Gromov-Wasserstein distance is discretized as the following quadratic assignment problem, 
\begin{equation*}
    \begin{gathered}
        GW^2(\bu_{\mX},D_{\mX},\bv_{\mY},D_{\mY}) = \min_{\Gamma\in S(\bu_{\mX},\bv_{\mY})} \mathcal{E}_{D_{\mX},D_{\mY}}(\Gamma),\\
     \mathcal{E}_{D_{\mX},D_{\mY}}(\Gamma)=\sum_{i,j}^M \sum_{p,q}^N (d^{\mX}_{ij} -d^{\mY}_{pq})^2 \gamma_{ip} \gamma_{jq},\\
    S(\bu_{\mX},\bv_{\mY})=\left\{  \Gamma= [\gamma_{ip}]_{M\times N}\  \Big| \sum_{i=1}^M  \gamma_{ip} = v_p, \sum_{p=1}^N  \gamma_{ip}= u_i,   \gamma_{ip} \geq 0, \  \forall{i,p}\right\}.
    \end{gathered}
\end{equation*}
It is non-convex and intractable to solve \cite{solomonEntropicMetricAlignment2016a}.

Entropic GW~\cite{peyreGromovWassersteinAveragingKernel2016} is proposed to minimize the GW objective with an entropic regularization term.
\begin{gather}\label{eq:entropic GW}
GW_{\veps}^2(\bu_{\mX},D_{\mX},\bv_{\mY},D_{\mY}) = \min_{\Gamma\in S(\bu_{\mX},\bv_{\mY})} \ \mathcal{E}_{D_{\mX},D_{\mY}}(\Gamma)+\veps H(\Gamma),\\
H(\Gamma)= \sum_{i=1}^M \sum_{p=1}^N \gamma_{ip} (\ln  \gamma_{ip}-1), \nonumber
\end{gather}
where $\veps > 0$ is the regularization parameter. 

\subsection{Mirror Descent Method}
We here introduce mirror decent \cite{beckMirrorDescentNonlinear2003,vishnoiAlgorithmsConvexOptimization2021}, which is the best-known method to solve Problem \eqref{eq:entropic GW}. Intuitively, it considers minimizing the linear approximation of the objective in iterations. To ensure a reliable linear approximation, a penalty function is simultaneously picked to prevent the new point from straying too far from the previous one. Taking KL divergence as the penalty term ultimately leads to the $l$-th iteration as
\begin{gather}
    \Gamma^{(l+1)} =  \argmin_{ \Gamma \in S(\bu_{\mX},\bv_{\mY})} \langle \nabla \mathcal{E}_{D_{\mX},D_{\mY}}( \Gamma^{(l)})+\veps \nabla H(\Gamma^{(l)}), \, \Gamma \rangle+\tau \text{KL}(\Gamma|\Gamma^{(l)}),\label{eq:GW proximal gradient} \\
     \text{KL}(\Gamma|\Gamma^{(l)}) = \sum_{i}^M \sum_{p}^N \left(\gamma_{ip} \ln{(\gamma_{ip}/\gamma_{ip}^{(l)})-\gamma_{ip}+\gamma_{ip}^{(l)}}\right), \nonumber
\end{gather}
where $\tau>0$ is the penalty coefficient. 
Rearranging equation~\eqref{eq:GW proximal gradient} gives that
\begin{equation}\label{eq:GW sinkhorn}
    \Gamma^{(l+1)} =\argmin_{ \Gamma \in S(\bu_{\mX},\bv_{\mY})} \langle \nabla \mathcal{E}_{D_{\mX}, D_{\mY}}(\Gamma^{(l)})+ (\veps - \tau) \nabla H(\Gamma^{(l)}), \, \Gamma \rangle+\tau H(\Gamma).
\end{equation}
It is an entropic Optimal Transport problem, with 
\begin{equation*}
    \Pi=\nabla \mathcal{E}_{D_{\mX}, D_{\mY}}(\Gamma^{(l)})+ (\veps - \tau) \nabla H(\Gamma^{(l)})
\end{equation*}
being cost matrix, which can be solved with the Sinkhorn algorithm \cite{cuturiSinkhornDistancesLightspeed2013a}.

As we stated before, the evaluation of $\Pi$ is time-consuming. More concretely, evaluating its first term $\nabla \mathcal{E}_{D_{\mX},D_{\mY}} (\Gamma)$ requires $O(M^2N^2)$ time for the reason that its $(i,p)$-th entry reads 
\begin{equation}\label{eq:gradient entry}
     \left[\nabla \mathcal{E}_{D_{\mX},D_{\mY}}(\Gamma)\right]_{ip}= 2\sum_{j=1}^M \sum_{q=1}^N (d^{\mX}_{ij} -d^{\mY}_{pq})^2 \gamma_{jq}.
\end{equation}
That is far more than the $O(MN)$ time of the Sinkhorn algorithm. Fortunately, it is observed in ~\cite{peyreGromovWassersteinAveragingKernel2016} that $\nabla \mathcal{E}_{D_{\mX},D_{\mY}}(\Gamma)$ can be decomposed into a constant term $\mathcal{C}_1$ and a linear term of $T$. Specifically,
\begin{gather*}
     \nabla \mathcal{E}_{D_{\mX},D_{\mY}}(\Gamma) = \mathcal{C}_1 - 4D_{\mX} \Gamma D_{\mY},\\
     \mathcal{C}_1=2((D_{\mX} \odot D_{\mX})
     \bu_{\mX}\mathbf{1}_N^\top +(D_{\mY}\odot D_{\mY})\bu_{\mY}\mathbf{1}_M^\top),
\end{gather*}
where $\odot$ is the Hadamard product. Computing $\mathcal{C}_1$ costs $O(M^2+N^2+MN)$ time and would be only performed once. Therefore, with the decomposition, the overall complexity of the solution of Entropic GW is reduced to $O(MN^2+M^2N)$, dominated by the computation of $D_{\mX} \Gamma D_{\mY}$. This complexity is still unacceptable in practice and worth our further exploration. 
\begin{remark}
    In the existing literature, $\tau = \veps$ is suggested~\cite{peyreGromovWassersteinAveragingKernel2016}. Then equation \eqref{eq:GW sinkhorn} turns to
    \begin{equation*}
     \Gamma^{(l+1)} \leftarrow \argmin_{\Gamma \in S(\bu_{\mX},\bv_{\mY})} \langle \nabla \mathcal{E}_{D_{\mX},D_{\mY}}(\Gamma^{(l)}), \Gamma \rangle+\veps H(\Gamma).
    \end{equation*}
We follow this in the subsequent discussion. Since it takes only $O(MN)$ time to calculate $\nabla H(\Gamma^{(l)})$, whether $\tau$ equals $\veps$ makes no difference in our statement about complexity.
\end{remark}
\begin{remark}\label{remark:FGW}
    (Entropic algorithm for FGW) Let $C=[c_{ip}]_{M\times N}$ be an additional cost matrix between $\bu_\mX$ and $\bv_\mY$. 
Fused GW (FGW) minimizes the objective  
\begin{equation*}
    \bar{\mathcal{E}}_{D_{\mX},D_{\mY}}(\Gamma)=(1-\theta)\cdot \sum_{i=1}^M \sum_{p=1}^N c_{ip}^2\gamma_{ip}+\theta \cdot \sum_{i,j}^M \sum_{p,q}^N (d^{\mX}_{ij} -d^{\mY}_{pq})^2 \gamma_{ip} \gamma_{jq}.
\end{equation*}
in the region $S(\bu_{\mX},\bv_{\mY})$, where $\theta \in [0,1]$ balances the effect of the linear and quadratic assignment \cite{vayerFusedGromovWassersteinDistance2020a}. The iteration formula \eqref{eq:GW sinkhorn} applies to FGW as well. Analogous to GW, its gradient has the decomposition
\begin{gather*}
     \nabla \bar{\mathcal{E}}_{D_{\mX},D_{\mY}}(\Gamma) = \mathcal{C}_2 - 4\theta \cdot D_{\mX} \Gamma D_{\mY},\\
     \mathcal{C}_2=(1-\theta)\cdot C\odot C+2\theta \cdot ((D_{\mX} \odot D_{\mX})
     \bu_{\mX}\mathbf{1}_N^\top +(D_{\mY}\odot D_{\mY})\bu_{\mY}\mathbf{1}_M^\top).
\end{gather*}
$C_2$ can also be computed in $O(M^2+N^2+MN)$ time, so $D_{\mX} \Gamma D_{\mY}$ still dominates the overall complexity.
\end{remark}
\begin{remark}
    (Entropic algorithm for UGW) Given $\rho>0,\hat{S}=\{\Gamma| \gamma_{ip}\geq 0, \forall{i,p}\}$, the Unbalanced Gromov-Wasserstein divergence is defined as
\begin{gather*}
    \min_{\Gamma \in \hat{S}} \mathcal{E}_{D_{\mX},D_{\mY}}(\Gamma)+\rho KL\left((\Gamma\mathbf{1})\otimes (\Gamma \mathbf{1})|\bu_\mX \otimes \bu_\mX\right)+\rho KL\left((\Gamma^\top \mathbf{1}) \otimes (\Gamma^\top \mathbf{1})|\bv_\mY\otimes \bu_\mY)\right).
\end{gather*}
The key to its entropic algorithms is to solve
\begin{align*}
    \min_{\Gamma \in \hat{S}} \langle \frac{1}{2}\nabla &\mathcal{E}_{D_{\mX},D_{\mY}}(\hat{\Gamma})+g(\hat{\Gamma}), \Gamma \rangle\\
          &+\mathbf{1}^\top \hat{\Gamma}\mathbf{1}\left(\rho \text{KL}(\Gamma \mathbf{1}|\bu_\mX)+\rho \text{KL}(\Gamma^\top \mathbf{1}|\bu_\mY)+\veps \text{KL}(\Gamma|\bu_\mX \otimes \bu_\mY)\right)
\end{align*}
    for an updated $\hat{\Gamma} \in \hat{S}$ at each iteration \cite{sejourneUnbalancedGromovWasserstein2021}. 
    Still the computation of $\nabla\mathcal{E}_{D_{\mX},D_{\mY}}(\hat{\Gamma})$ or rather $D_{\mX} \hat{\Gamma} D_{\mY}$ is to blame for $O(M^2N+MN^2)$ complexity, while the other parts take no more than $O(M^2+N^2+MN)$ time. The method proposed in this paper will also apply here.
\end{remark}

\section{Fast Gradient Computation}\label{sec:3}

In this section, we present an efficient method that computes $D_\mX\Gamma D_\mY$ in $O(MN)$ time. For the sake of simplicity, we assume $M=N$. \footnote{The method can easily handle the case where $M$ is not equal to $N$. This assumption is without loss of generality.} 

First, we note that the distance matrices $D_\mX, D_\mY$ satisfy
\begin{equation*}
       D_\mX = h_{\mX}^k \tilde{D}, \quad D_\mY = h_{\mY}^k \tilde{D},
\end{equation*}
where 
\begin{equation*}
\tilde{D} = L+L^\top, \quad
L=
\begin{pmatrix}
    0 & 0 & \cdots & & 0\\
    1 & 0 &\cdots & & 0\\
    2 & 1 &\ddots &  & \vdots\\
    \vdots & \vdots  & \ddots &\ddots & \\
    (N-1) & (N-2) &\cdots & 1  & 0
\end{pmatrix}^{\odot k}.
\end{equation*}
The operator $\odot k$ refers to raising each element of the matrix to the power of $k$.
Therefore $D_\mX \Gamma D_\mY$ can be expanded as
\begin{multline}\label{eq:Dx}
        D_\mX \Gamma D_\mY  = h_{\mX}^k h_{\mY}^k (\tilde{D} \Gamma \tilde{D})
        =h_{\mX}^k h_{\mY}^k\left(L \Gamma L+L \Gamma L^\top+L^\top \Gamma L+L^\top \Gamma L^\top\right)\\
    =h_{\mX}^k h_{\mY}^k\left(L  \left(L^\top\Gamma^\top\right)^\top+L  \left(L\Gamma^\top\right)^\top+L^\top \left(L^\top\Gamma^\top\right)^\top+ L^\top\left(L\Gamma^\top\right)^\top\right).
\end{multline}

Next, we show that for any $\bx=(x_1,x_2,\cdots x_N)^\top \in \mathbb{R}^{N}$,  $\by = L\bx$ can be implemented with $O(N)$ element-wise operations. $L^T\bx$ can be computed in a similar way. As a result, the total time cost of computing equation \eqref{eq:Dx} will be no more than $O(N^2)$.
To be concrete,
\begin{equation}\label{1d y}
    \by = \begin{pmatrix}
        0,\sum_{j=1}^{1} (2-j)^{k} x_j, \cdots, \sum_{j=1}^{N-1} (N-j)^{k} x_j 
    \end{pmatrix}^{\top}.
\end{equation}
Defining $N \times (k+1)$ elements \begin{equation*}
    a_{ir} = \sum_{j=1}^{i-1} (i-j)^{r-1} x_j, \quad i \in \{1,\cdots N\},\quad r\in \{1,\cdots,k+1\},
\end{equation*} 
we obtain that the $i$-th entry of $\by$
\begin{equation*}
    y_i = a_{i,k+1},\quad i \in \{1,\cdots N\}.
\end{equation*}
A significant observation is that 

\begin{multline}\label{eq:1d recursive relation}
     a_{i+1,r} =  x_{i}+\sum_{j=1}^{i-1} (i-j+1)^{r-1} x_j  = x_{i}+\sum_{j=1}^{i-1}\sum_{s=1}^{r}\binom{r-1}{s-1} (i-j)^{s-1} x_j
        \\= x_{i}+ \sum_{s=1}^{r}\binom{r-1}{s-1} \sum_{j=1}^{i-1}(i-j)^{s-1} x_j
        =x_{i}+\sum_{s=1}^{r}\binom{r-1}{s-1}a_{is}
         \\= x_{i}+a_{i1}+\binom{r-1}{1}a_{i2}+\cdots+\binom{r-1}{r-1}a_{ir}.
\end{multline}
In other word, we can calculate $a_{i+1,r}$ by making linear combination of $\{a_{is}\}_{s=1}^r$ recursively. With $\{a_{is}\}_{s=1}^r$ and all the binomial coefficients being known\footnote{In fact, all the binomial coefficients can be computed in $O(k^2)$ time \cite{rolfeBinomialCoefficientRecursion2001}.}, it needs only $r-1$ multiplications and $r$ additions to get $a_{i+1,r}$. 

In view of the fact that $a_{1r} = 0$ for any $r$, we conclude that it takes
\begin{equation*}
\quad (N-1)\sum_{r=1}^{k+1}(r-1)=(N-1) \frac{k(k+1)}{2}
\end{equation*}
multiplications and 
\begin{equation*}
\quad (N-1)\sum_{r=1}^{k+1}r=(N-1) \frac{(k+2)(k+1)}{2}
\end{equation*}
additions
to evaluate all $a_{ir}$ in the order of 
\begin{equation*}
    a_{11},\cdots a_{N,1},\cdots,a_{1,k+1},\cdots a_{N,k+1}.
\end{equation*}
Therefore, the computation of $\by=L\bx$ is finished in $O(k^2N)$ time. In practice, $k$ refers to the power of the distance and typically takes $1$ or $2$. So the total cost is $O(N)$ for short.

 \subsection{Extension to High Dimension}~\label{sec:4}
In this part, we illustrate the generalization of FGC-GW to 2D space. And there is no essential difference to generalizing the algorithm to higher dimensional space.

Consider two probabilistic distributions:
\begin{equation*}
    \bu_{\mX}=
    \begin{pmatrix}
        u_{11} & u_{12} & \cdots & u_{1n}\\
        u_{21} & u_{22} & \cdots & u_{2n}\\
        \vdots & \vdots & \ddots & \vdots\\
        u_{n1} & u_{n2} & \cdots & u_{nn}
    \end{pmatrix}, \quad
    \bv_{\mX}=
    \begin{pmatrix}
        v_{11} & v_{12} & \cdots & v_{1n}\\
        v_{21} & v_{22} & \cdots & v_{2n}\\
        \vdots & \vdots & \ddots & \vdots\\
        v_{n1} & v_{n2} & \cdots & v_{nn}
    \end{pmatrix}
\end{equation*}
on two uniform 2D grids with both sizes $n\times n$, which contributes to the total number of grid points $N=n^2$. We set both the horizontal and vertical spacing of space $\mX$ to $h_{\mX}$ and those of space $\mY$ to $h_{\mY}$. Flatting $\bu_{\mX}$ and $\bv_{\mY}$ into vectors in column-major order, by analogy with \eqref{1d distance matrices}, we get two distance matrices
where
\begin{equation}\label{eq:2d Dx}
    \hspace{-0.9mm}\;\,D_{\mX}=h_{\mX}^k\hat{D}, \quad
    D_{\mY} = h_{\mY}^k\hat{D}.
\end{equation}
Here
\begin{equation*}
    \hat{D}=\begin{pmatrix}
    D_1 & D_1+J & D_1+2J& \cdots &  D_1+(n-1)J\\
    D_1+J & D_1 & D_1+J& \cdots &  D_1+(n-2)J\\
    D_1+2J & D_1+J& D_1 &\cdots &  D_1+(n-3)J\\
    \vdots&\vdots & \vdots &\ddots & \vdots \\
    D_1+(n-1)J & D_1+(n-2)J &D_1+(n-3)J & \cdots & D_1
    \end{pmatrix}_{n^2\times n^2}^{\odot k}
\end{equation*}
with
\begin{equation*} D_1=\begin{pmatrix}
    0 & 1 & 2& \cdots & n-1\\
    1 & 0 & 1 &\cdots &  n-2\\
    2 & 1 & 0 &\cdots &  n-3\\
    \vdots&\vdots&\vdots & \ddots &\vdots \\
    n-1 & n-2& n-2&\cdots & 1
\end{pmatrix}_{n\times n}, \quad \quad J = \begin{pmatrix}
        1 & \cdots & 1\\
        \vdots  &\ddots & \vdots\\
        1 & \cdots & 1 
    \end{pmatrix}_{n\times n}.
\end{equation*}
At this point,
\begin{equation}\label{2d gradient}
        D_{\mX}\Gamma D_{\mY} = h_{\mX}^k h_{\mY}^k (\hat{D} \Gamma \hat{D})
        =h_{\mX}^k h_{\mY}^k\left(\hat{D}(\hat{D}\Gamma^\top)^\top\right).
\end{equation}
Notice that any $\Gamma \in \mathbb{R}^{n^2 \times n^2}$ can be partitioned into $n^2$ vectors in $\mathbb{R}^{n^2}$.
Whereafter, we show that for any $x \in  \mathbb{R}^{n^2}$, $\hat{D} x$ can be performed in $O(n^2)$ time. In this way, \eqref{2d gradient} can be finished in total $O(N^2)=O(n^4)$ time. 

One can expand $\hat{D}$ as
\begin{equation*}
    \hat{D} = \sum_{r=0}^k \binom{k}{r}D_1^{\odot r}\otimes D_1^{\odot {k-r}},
\end{equation*}
with the notion $\otimes$ representing the Kronecker product.
Therefore
\begin{equation}\label{eq:2D kronecker}
    \hat{D}x = \sum_{r=0}^k \binom{k}{r}\left(D_1^{\odot r}\otimes D_1^{\odot {k-r}}\right)x = \sum_{r=0}^k \binom{k}{r}\text{vec}\left(D_1^{\odot k-r} \text{mat}(x) D_1^{\odot {r}}\right),
\end{equation}
where $\text{vec}(\cdot)$ denotes the vectorization of a matrix, that is, for $Q =[q_{ij}]_{n \times n} \in \mathbb{R}^{n \times n}$,
\begin{equation*}
\begin{gathered}
    \text{vec}(Q)  = (q_{11}, \cdots, q_{1n},q_{21}, \cdots, q_{2n},\cdots,q_{n1},\cdots,q_{nn})^\top \in \mathbb{R}^{n^2}.
\end{gathered}
\end{equation*}
And $\text{mat}(\cdot)$ denotes the matrixization of a vector, i.e., the inverse transformation of vectorization. Note the similarity between $D_1^{\odot k-r} \text{mat}(x) D_1^{\odot {r}}$ and $\tilde{D} \Gamma \tilde{D}$ in equation~\eqref{eq:Dx}. It is obvious that $D_1^{\odot k-r} \text{mat}(x) D_1^{\odot {r}}$ can be computed in $O(k^2n^2)$ time, by leveraging the approach in 1D case. Totally, the cost of computing $\hat{D}x$ in equation~\eqref{eq:2D kronecker} is $O(k^4n^2)$, for short, $O(n^2)$.

\section{Numerical Experiments}~\label{sec:5}
In this section, we perform several experiments to validate the effectiveness and efficiency of the proposed method. We conduct experiments over the 1D and 2D random distributions, time series\cite{vayerFusedGromovWassersteinDistance2020a}, and images \cite{mnist, horse}. For the experiments on random distributions, we evaluate both Gromov-Wasserstein and Fused Gromov-Wasserstein (FGW) metrics. We compare the FGC with the original computation by Entropic (Fused) Gromov-Wasserstein.
For the alignment of time series \cite{vayerFusedGromovWassersteinDistance2020a} and image data \cite{mnist, horse}, we consider the FGW metric as it allows for the inclusion of feature information in addition to structure information. This characteristic renders FGW more suitable for alignment tasks in comparison to GW \cite{vayerFusedGromovWassersteinDistance2020a}.
 
To ensure the fairness of evaluation, we consider sequential implementation in this paper. The original entropic (F)GW and FGC implementations are realized using the C++ language, leveraging the vector inner-product functionality offered by the Eigen library. All experiments are executed with a single-thread program over a platform with 128G RAM and one Intel(R) Xeon(R) Gold 5117 CPU @2.00GHz.

\subsection{1D random distributions}
We consider two 1D random distributions $u_{\mX}=(u_1,\cdots,u_N)$ and $v_{\mY}=(v_1,\cdots,v_N)$ on uniform grid points 
\begin{equation*}
    x_i = y_i= \frac{i-1}{N-1}, \quad i=1,2,\cdots,N.
\end{equation*}
All $u_i$ and $v_i$ are sampled from a uniform distribution over $[0,1]$ and then normalized so that $u_{\mX}$ and $v_{\mY}$ are distributions. 
Our objective is to compare the performance and computational cost of our FGC algorithms and the original entropic algorithms on computing the corresponding GW metric and FGW metric ($\theta=0.5$). We take $k=1$ for $D_{\mX},D_{\mY}$ in \eqref{1d distance matrices} and $c_{ip}=|i-p|$ for $C=[c_{ip}]$ in Remark \ref{remark:FGW}. The number of iterations of mirror descent \eqref{eq:GW sinkhorn} is set to 10. 
 We test $100$ random experiments for each $N$.
\begin{table}[t]
\small
\centering
	\begin{tabular}{cclccc}
		\toprule
		\multirow{2}{*}{Metric} &
		\multirow{2}{*}N &
		\multicolumn{2}{c}{Computational time (s)} & \multirow{2}{*}{Speed-up ratio} & \multirow{2}{*}{$||P_{Fa}-P||_F$} \\
		
		\cline{3-4} & & \quad FGC &Original \\
		
		 \midrule
		\multirow{4}{*}{GW}
		 &500   & $4.97\times10^{-1}$ & $4.40\times10^{0}$ &
		 $8.85$  & $5.12\times10^{-15}$   \\ 
		&1000  & $2.13\times10^{0}$ & $3.46\times10^{1}$  & $16.2$ &$4.33\times10^{-15}$   \\ 
		&2000  & $1.01\times10^{1}$ & $2.80\times10^{2}$ & $27.7$ &$2.87\times10^{-15}$   \\ 
		&4000  & $5.01\times10^{1}$ & $2.44\times10^{3}$ & $48.7$ &$2.04\times10^{-15}$   \\ 
		 \midrule
		\multirow{4}{*}{FGW}
		 &500   & $6.00\times10^{-1}$ & $4.58\times10^{0}$ &
		 $7.63$  & $1.08\times10^{-15}$   \\ 
		&1000  & $2.54\times10^{0}$ & $3.53\times10^{1}$  & $13.9$ &$8.27\times10^{-16}$   \\ 
		&2000  & $1.20\times10^{1}$ & $2.83\times10^{2}$ & $23.6$ &$5.78\times10^{-16}$   \\ 
		&4000  & $5.73\times10^{1}$ & $2.47\times10^{3}$ & $43.1$ &$4.11\times10^{-16}$   \\ 
		\bottomrule
	\end{tabular}
 	\caption{1D random distribution. The comparison between the algorithms with FGC and the original ones with the different number of grid points $N$. For GW and FGW, the regularization parameters $\veps=0.002$, respectively. Column for $||P_{Fa}-P||_F$ validates the correctness of the results by FGC.}
	\label{tab:1D random}
\end{table}

In Table \ref{tab:1D random}, we show the averaged running time of the algorithms and the difference in the transport plans. We can see that FGC has an overwhelming advantage in computational
speed. Moreover, it produces almost identical transport plans as the original entropic algorithms.

\begin{figure}[t]
     \centering
     \begin{subfigure}[b]{0.45\textwidth}
         \centering
         \includegraphics[width=\textwidth]{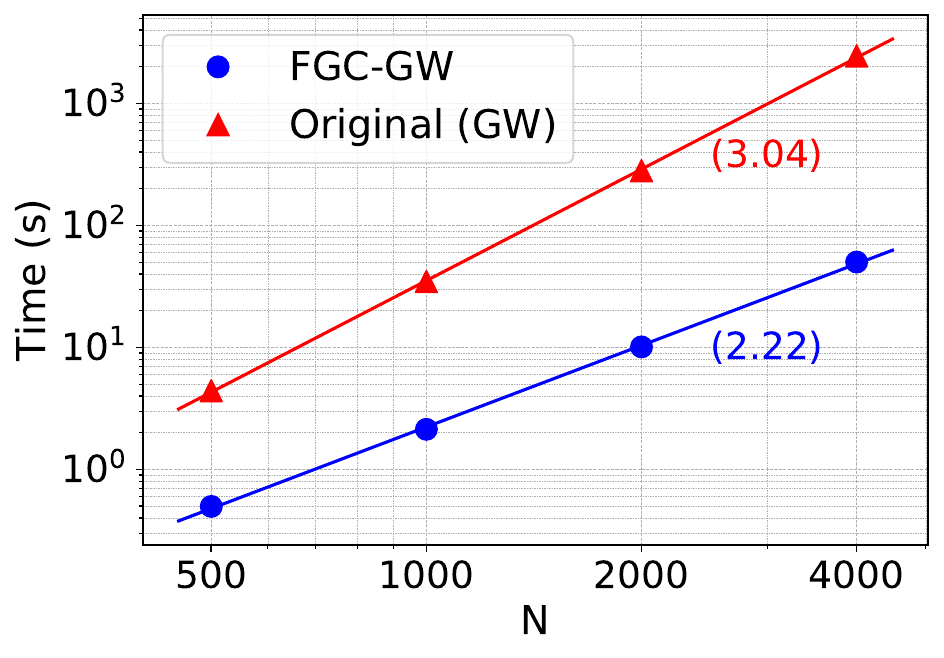}
     \end{subfigure}
     \begin{subfigure}[b]{0.45\textwidth}
         \centering
         \includegraphics[width=\textwidth]{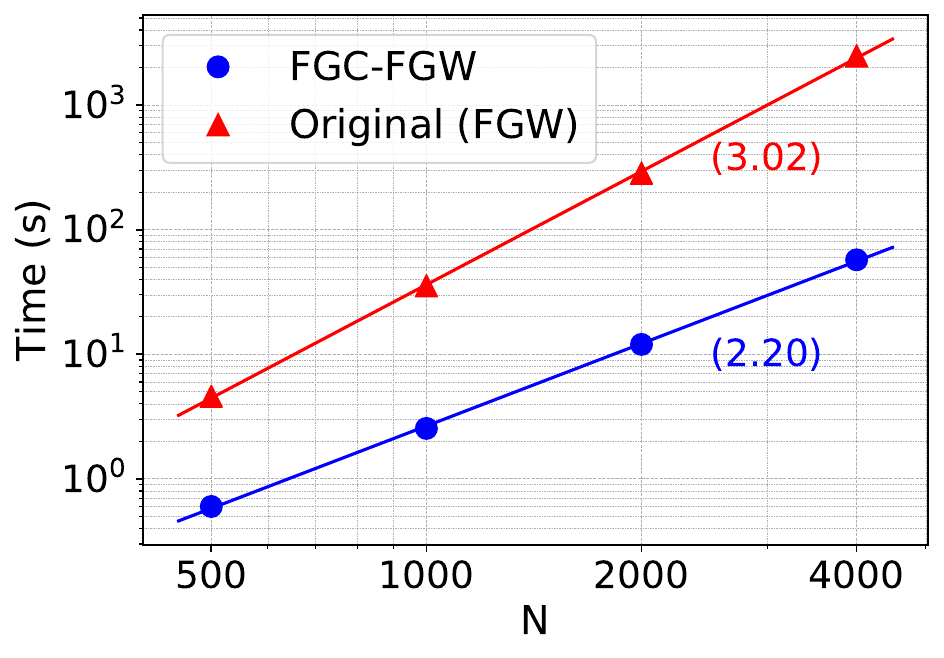}
     \end{subfigure}
        \caption{1D random distributions. The computation time for GW (\textbf{left}) and FGW (\textbf{right}) under different N. The numbers are the fitted slopes, representing the empirical computational complexities.}
     \label{fig:1d random}
\end{figure}

To study the efficiency advantage of FGC further, we visualize the relationship between time cost and the number of grid points in Figure~\ref{fig:1d random}. Note that the two axes are log scales. By data fitting, we find empirical $O(N^{2.22})$ and $O(N^{2.20})$ complexities of FGC on GW and FGW while those of the original algorithms are $O(N^{3.04})$and $O(N^{3.02})$.
\begin{table}[t]
\small
\centering
\begin{tabular}{cccccc}
    \toprule
    \multirow{2}{*}{Metric} &
    \multirow{2}{*}{$N=n\times n$} &
    \multicolumn{2}{c}{Computational time (s)} & \multirow{2}{*}{Speed-up ratio} & \multirow{2}{*}{$||P_{Fa}-P||_F$} \\
    
    \cline{3-4} & & FGC &	Original \\
    
     \midrule
    \multirow{4}{*}{GW}
     &$30 \times 30$   & $1.73\times10^{0}$ & $2.46\times10^{1}$ &
     $14.2$  & $3.03\times10^{-14}$   \\ 
    &$60 \times 60$  & $5.53\times10^{1}$ & $1.66\times10^{3}$  & $30.0$ &$7.94\times10^{-15}$   \\ 
    &$90 \times 90$  & $3.01\times10^{2}$ & $1.85\times10^{4}$ & $61.5$ &$6.75\times10^{-15}$   \\ 
    &$120 \times 120$  & $9.65 \times10^{2}$& $-$ & $-$ &$-$   \\ 
     \midrule
    \multirow{4}{*}{FGW}
 &$30 \times 30$   & $1.84\times10^{0}$ & $2.50\times10^{1}$ &
     $13.6$  & $2.56\times10^{-14}$   \\ 
    &$60 \times 60$  & $5.90\times10^{1}$ & $1.71\times10^{3}$  & $29.0$ &$1.48\times10^{-15}$   \\ 
    &$90 \times 90$  & $3.22\times10^{2}$ & $1.89\times10^{4}$ & $58.7$ &$1.00\times10^{-15}$   \\ 
    &$120 \times 120$  & $1.08\times10^{3}$ & $-$ & $-$ &$-$   \\ 
    \bottomrule
\end{tabular}
\caption{2D random distributions. The comparison between the algorithms with FGC and the original ones. For GW and FGW, the regularization parameters $\veps=0.004$, respectively. Column for $||P_{Fa}-P||_F$ validates the correctness of the results by FGC. A dash means the running time exceeds 10 hours.}
\label{tab:2d random}
\end{table}

\begin{figure}[t]
     \centering
     \begin{subfigure}[c]{0.45\textwidth}
         \centering
         \includegraphics[width=\textwidth]{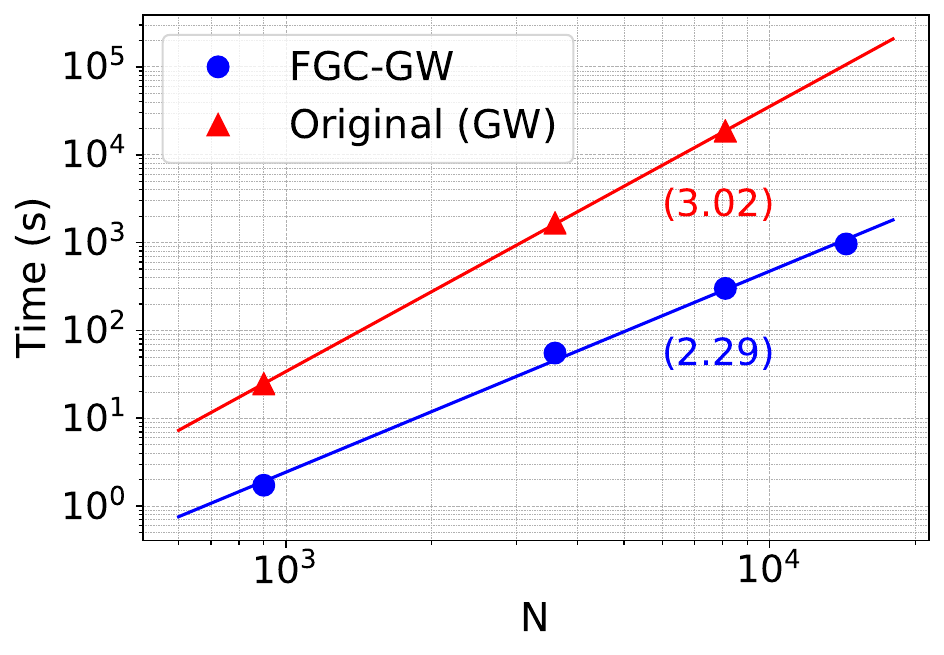}
     \end{subfigure}
     \begin{subfigure}[c]{0.45\textwidth}
         \centering
         \includegraphics[width=\textwidth]{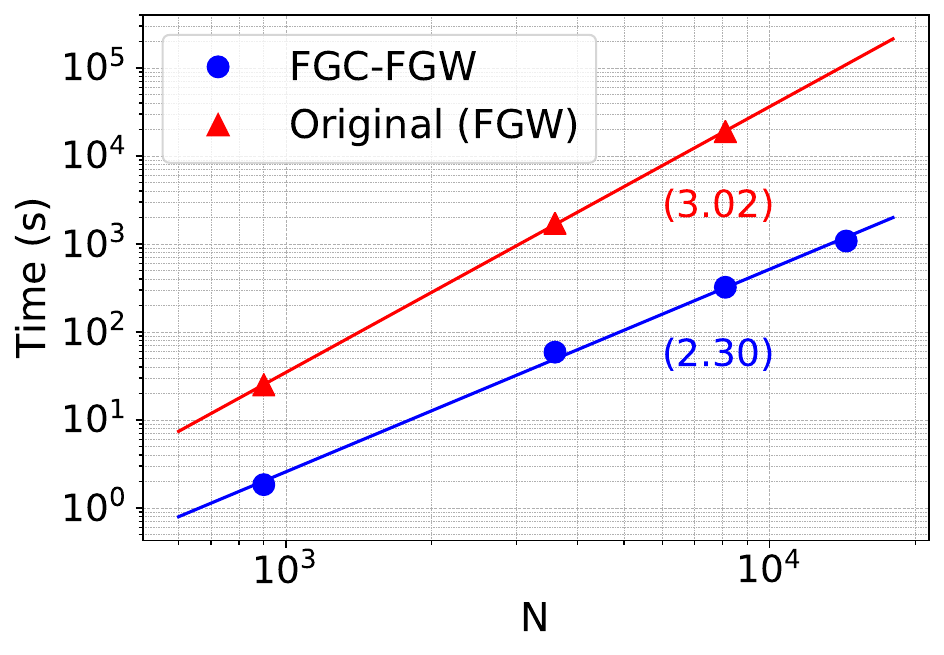}
     \end{subfigure}
     \caption{2D random distributions. The computation time for GW (left) and FGW (right) under different $N$. The numbers are the fitted slopes, representing the empirical computational complexities.}
     \label{fig:2d random}
\end{figure}

\subsection{2D random distributions}
Next, we investigate the performance of FGC on 2D random distributions. The supports of distributions are $N=n \times n$ grid points uniform scattered on $[0,1]\times[0,1]$. 

Table \ref{tab:2d random} and Figure \ref{fig:2d random} show the results. It can be seen that the fast algorithms keep the computational advantage and make large-scale tasks possible. We also estimate their empirical complexities, $O(N^{2.29})$ for GW and $O(N^{2.30})$ for FGW while those of the original versions are $O(N^{3.02})$ and $O(N^{3.02})$, respectively.

\subsection{Time series alignment}
Time series data is widely generated on a daily basis in various application domains, including bioinformatics \cite{bar-josephAnalyzingTimeSeries2004,bar-josephStudyingModellingDynamic2012}, finance \cite{majumdarClusteringClassificationTime2020,arroyoDifferentApproachesForecast2011}, engineering \cite{pastenTimeSeriesAnalysis2018,mudelseeTrendAnalysisClimate2019}, and others. Classifying different types of time series data has emerged as a significant focus in machine learning research in recent years \cite{abandaReviewDistanceBased2019}. As a pre-work, it is highly important to find a good similarity measure for time series data  \cite{sejourneUnbalancedGromovWasserstein2021,abandaReviewDistanceBased2019}. Superior to the GW metric, FGW can effectively incorporate both signal strength (linear term) and time information (quadratic term), enabling comparatively accurate alignment of time series waveforms. Consequently, it is considered more appropriate for defining time series similarity \cite{vayerFusedGromovWassersteinDistance2020a}.

We here investigate the acceleration effect of FGC on the FGW metric in the time series alignment task. Consider a series in [0,1] that consists of two humps with heights of 0.5 and 0.8. We construct the other series by moving the humps around. Now we would like to align them using the transport plan of FGW by setting $k=1$ for $D_\mX$ and $D_\mY$ in equation~\eqref{1d distance matrices} and $C$ as the signal strength difference. After uniform dividing the time axis, we get $N$ sampling points for each series and then compute FGW ($\theta=0.5$) with two algorithms. Likewise, we repeat the experiment 100 times and record the time and plans.
\begin{table}[t]
\small
\centering
	\begin{tabular}{clccc}
		\toprule
		\multirow{2}{*}N &
		\multicolumn{2}{c}{Computational time (s)} & \multirow{2}{*}{Speed-up ratio} & \multirow{2}{*}{$||P_{Fa}-P||_F$} \\
		
		\cline{2-3}  & FGC-FGW &	original \\
		
		 \hline
		 400   & $3.78\times10^{-1}$ & $2.43\times10^{0}$ & $6.43$  & $2.02\times10^{-15}$   \\ 
		800  & $1.59\times10^{0}$ & $1.91\times10^{1}$  & $12.0$ &$1.45\times10^{-15}$   \\ 
		1600  & $7.02\times10^{0}$ & $1.54\times10^{2}$ & $21.9$ &$1.08\times10^{-15}$   \\ 
		3200  & $3.59\times10^{1}$ & $1.24\times10^{3}$ & $34.5$ &$7.17\times10^{-16}$   \\
		\bottomrule
	\end{tabular}
	\caption{Time series tasks with FGW metric. The comparison between the fasts algorithms with FGC and the original ones with the different number of grid points $N$. Column for $||P_{Fa}-P||_F$ validates the correctness of the results by FGC.}
	\label{tab:time series}
\end{table}

\begin{figure}[htbp]
\centering
\begin{minipage}[c]{0.45\textwidth}
\centering
\vspace{2.5mm}
\includegraphics[width=\textwidth]{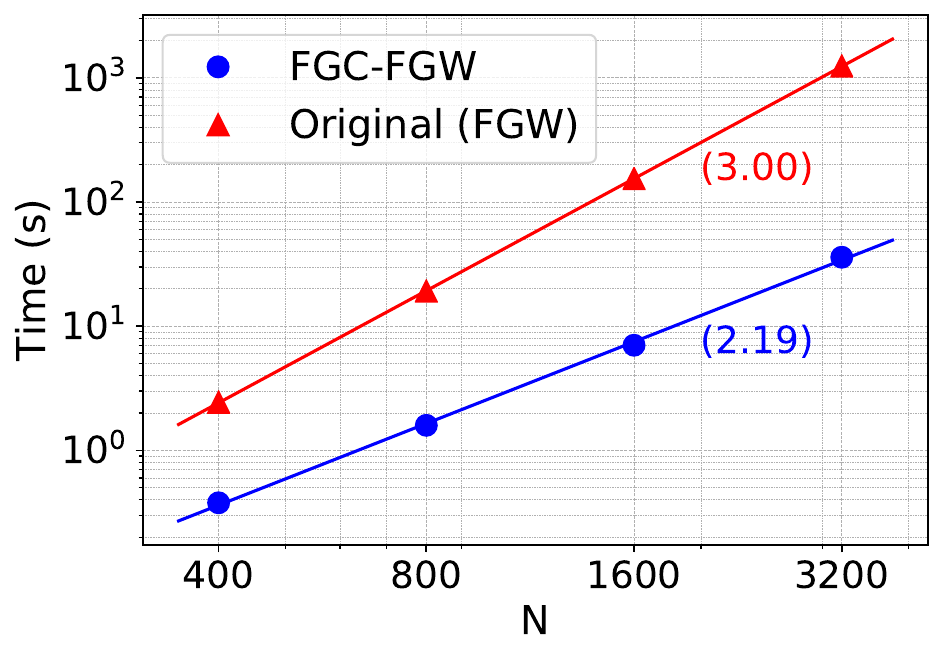}
\end{minipage}
\begin{minipage}[c]{0.45\textwidth}
\centering
\includegraphics[width=\textwidth]{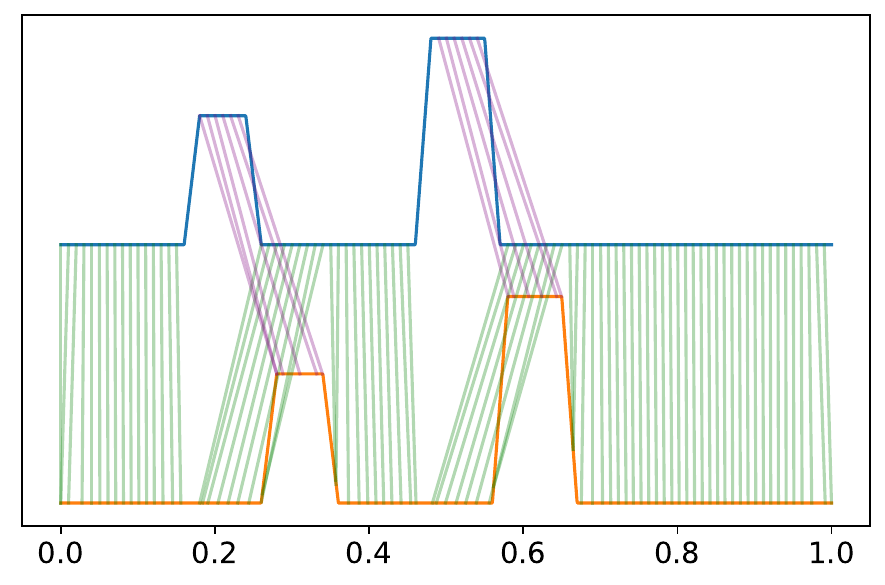}
\end{minipage}
     \caption{Time series alignment with FGW metric. \textbf{Left}: The time comparison on computing FGW metric for different numbers of sample points with entropic GW with or without FGC. \textbf{Right}: The visualized alignment between source (Blue) and target (Orange) time series at $N=1600$. The lines across two time series represent the transport plan. }
     \label{fig:time series}
\end{figure}

Table~\ref{tab:time series} reports the average results and Figure~\ref{fig:time series} presents
the $O(N^{2.19})$ empirical complexity of FGC on the left. As expected, the fast algorithm demonstrates a clear computational speed advantage. Moreover, we give the plan at $N=200$ explicitly on the right of Figure~\ref{fig:time series}, illustrating the application value of FGW for time series classification tasks.

\subsection{Image alignment}
Measuring the similarity between images is another meaningful application of FGW \cite{vayerFusedGromovWassersteinDistance2020a}. Intuitively, good alignment between images implies a quite accurate depiction of similarity \cite{vayerFusedGromovWassersteinDistance2020a}. Likewise, FGW does a better job than GW in this task, because it can take advantage of not only pixel coordinates (quadratic term) but also pixel values (linear term). Nevertheless, suffering from high computational complexity, FGW is prevented from applying to high-resolution images. By reducing the computational time significantly, FGC makes it possible.
\subsubsection{Three invariances on handwritten digits image data}
FGW showcases that the alignment can be invariant under types of transformations, such as translation, rotation, and reflection. We will show that FGC acceleration preserves the exact same invariance.
We first selected one $28 \times 28$ image representing digits $3$ from MNIST dataset \cite{mnist}, then we made the new ones by translation, mirroring, and rotation of the original image. The objective is to align the original image with the others. We use the Manhattan distance on the pixel coordinate gird for $D_{\mX}$ and $D_{\mY}$, i.e., take $k=1,\; h_\mX=h_\mY=1$ in equation~\eqref{eq:2d Dx}. $C$ is constructed by calculating the difference in the pixel gray levels between source pixels and target pixels. Referring to \cite{vayerFusedGromovWassersteinDistance2020a}, we take $\theta=0.1$. 

\begin{figure}[t]
     \centering
     \begin{subfigure}[c]{0.3\textwidth}
         \centering
         \includegraphics[width=\textwidth]{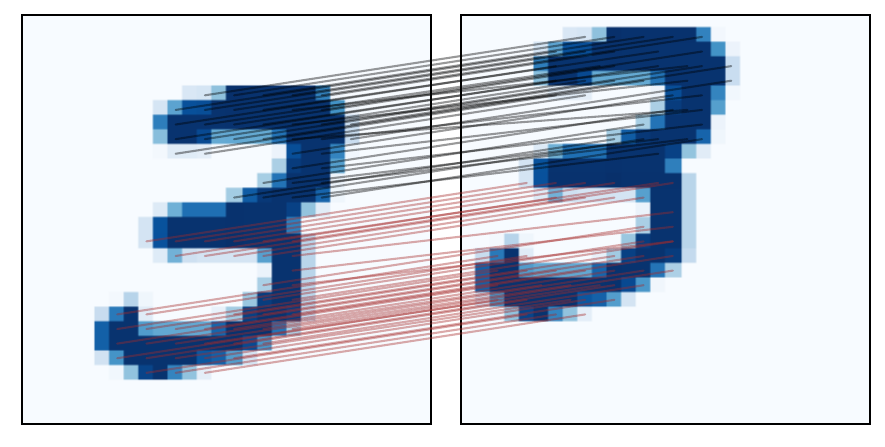}
     \end{subfigure}
     \begin{subfigure}[c]{0.3\textwidth}
         \centering
         \includegraphics[width=\textwidth]{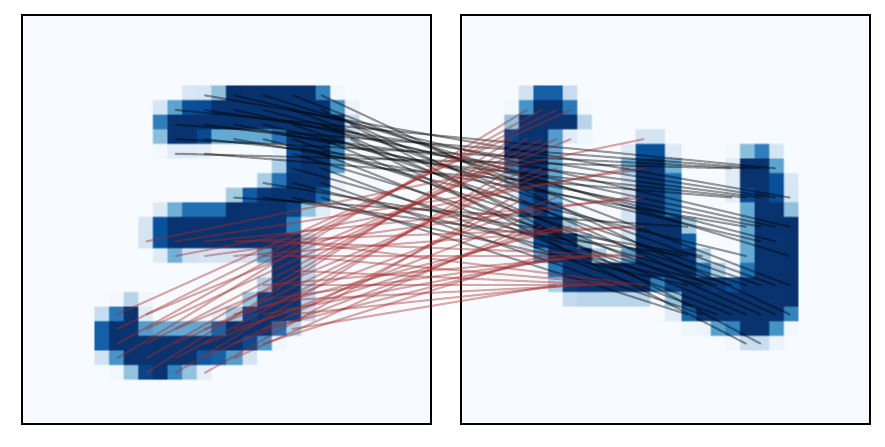}
     \end{subfigure}
    \begin{subfigure}[c]{0.3\textwidth}
         \centering
         \includegraphics[width=\textwidth]{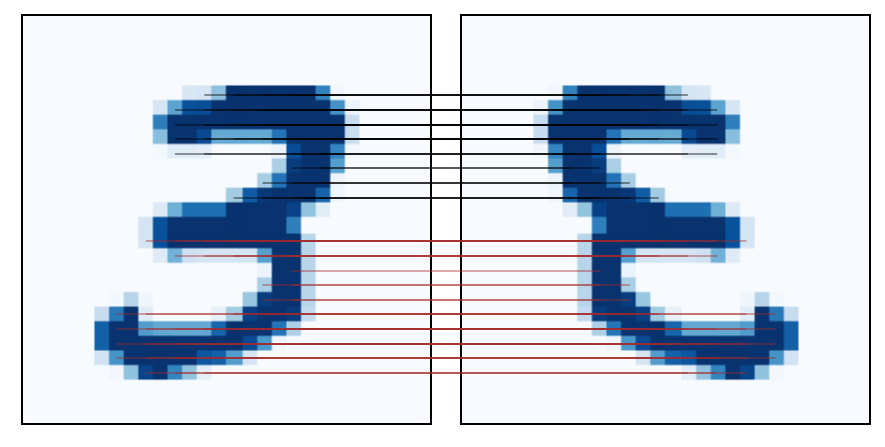}
     \end{subfigure}
     \caption{Handwritten digits task with FGW metric. The original image is matched to the new ones made by translation (\textbf{left}), rotation (\textbf{mid}), and reflection (\textbf{right}). The black lines and red lines represent transport plans.}
     \label{fig:digit}
\end{figure}

The average results of 100 runs are shown in Table~\ref{tab:digit}. It is observed that our fast algorithm defeats the original one again, which costs about $3$s and achieves about 10 times the speedup. Figure~\ref{fig:digit} shows the actual transport plans. For a clear view, plans are marked with two colors. Graphically, the FGW metric makes each pixel of the digit aligned well.

\begin{table}[t]
\small
\centering
	\begin{tabular}{llccc}
		\toprule
		\multirow{2}{*}{Invariance} &
		\multicolumn{2}{c}{Computational time (s)} & \multirow{2}{*}{Speed-up ratio} & \multirow{2}{*}{$||P_{Fa}-P||_F$} \\
		\cline{2-3}  & FGC-FGW &	Original \\
		 \midrule
		 Translation   & $2.86\times10^{0}$ & $2.86\times10^{1}$ & $10.0$  & $6.96\times10^{-14}$   \\ 
		Rotation  & $2.34\times10^{0}$ & $2.26\times10^{1}$  & $9.66$ &$1.51\times10^{-14}$   \\ 
		 Reflection  & $2.34\times10^{0}$ & $2.27\times10^{1}$ & $9.70$ &$1.39\times10^{-14}$   \\
		\bottomrule
	\end{tabular}
	\caption{Handwritten digits task with FGW metric. The comparison between FGC\-FGW and the original algorithm on aligning three pairs of images under different types of transformations.}
	\label{tab:digit}
\end{table}

\subsubsection{Complex deformation on horse image data}
In this part, to further emphasize the practical value of FGC on images, we conduct large-scale experiments on two $450 \times 300$ images of a running horse captured from a video \cite{horse}. The shapes of the horse in the images reveal complex deformation during running. Before alignment, they are subsampled to $N = n\times n$ first and then converted to grayscale images. We set $D_\mX, D_\mY$ and $C$ the same as the handwritten digits task, except for taking $h_\mX = h_\mY = 100/n$ to make $D_\mX$ and $D_\mY$ comparable with $C$ in magnitude. We computed FGW with various $\theta = 0.4, 0.6, 0.8$ to show that FGC captures the original invariance with complexity advantage.

\begin{figure}[htbp]
\centering
\begin{minipage}[c]{0.45\textwidth}
\centering
\vspace{2.5mm}
\includegraphics[width=\textwidth]{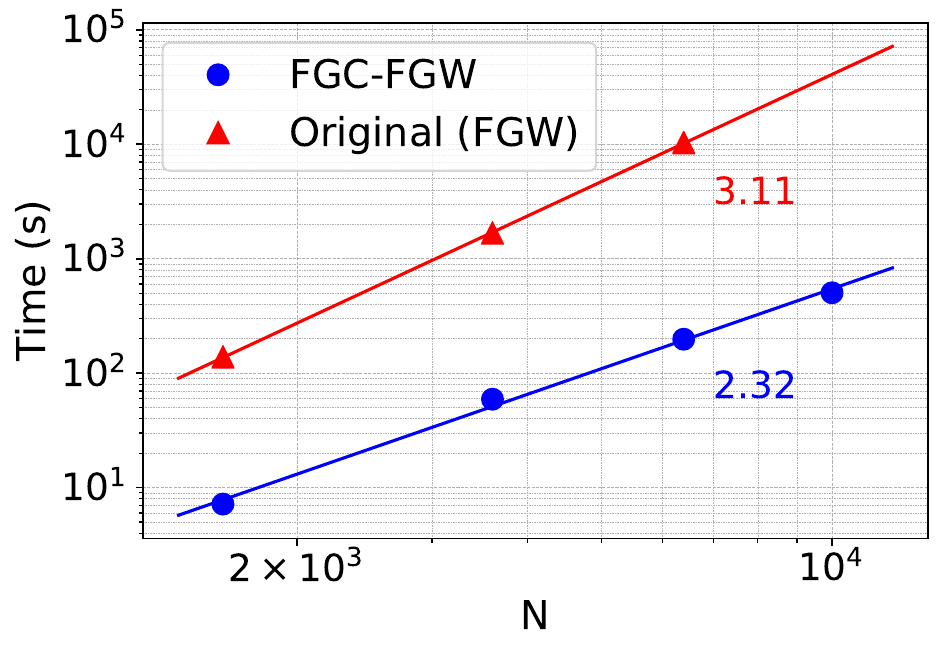}
\end{minipage}
\begin{minipage}[c]{0.53\textwidth}
\centering
\includegraphics[width=\textwidth]{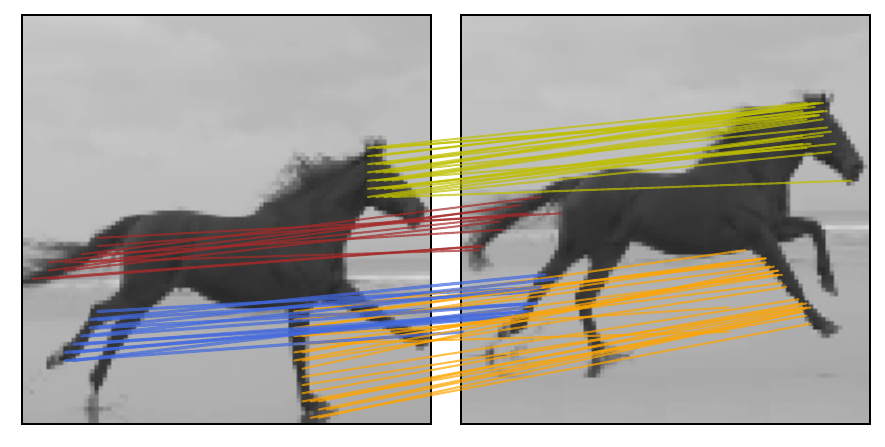}
\vspace{0.005mm}
\end{minipage}
     \caption{Horse images task with FGW metric. \textbf{Left}: the time comparison between FGC\-FGW and the original algorithm with $\theta=0.8$. The numbers attached are fitted slops, representing empirical complexities. \textbf{Right}: the transport plans with $\theta=$ 0.8 on $100\times 100$ images.}
     \label{fig:horse time plan}
\end{figure}

Table~\ref{tab:horse} reports the average results of $100$ runs and Figure~\ref{fig:horse time plan} presents the $O(N^{2.32})$ empirical complexity of FGC at $\theta=0.8$ on the left. FGC enables the processing of all images in less than 10 minutes regardless of $N$ while the original algorithm would struggle to deal with those of size $100 \times 100$. When $N=80$, FGC brings a 40x acceleration roughly. 
And this would be promoted to more than 60x at N=100. Figure~\ref{fig:horse time plan} also shows the plans obtained at $\theta=0.8,\; N=100$. We can see that the horse's head, tail, and legs are aligned well.  
\begin{table}[h]
\small
\centering
	\begin{tabular}{cccccc}
		\toprule
		\multirow{2}{*}{$\theta$} &
		\multirow{2}{*}{$N = n\times n$} &
		\multicolumn{2}{c}{Computational time (s)} & \multirow{2}{*}{Speed-up ratio} & \multirow{2}{*}{$||P_{Fa}-P||_F$} \\
		
		\cline{3-4} & & FGC-FGW &	Original \\
		 \midrule
		\multirow{4}{*}{0.4}
		 &$40 \times 40$   & $7.08\times10^{0}$ & $1.39\times10^{2}$ &
		 $19.6$  & $4.25\times10^{-16}$   \\ 
		&$60 \times 60$  & $5.92\times10^{1}$ & $1.68\times10^{3}$  & $28.4$ &$3.54\times10^{-16}$   \\ 
		&$80 \times 80$  & $1.99\times10^{2}$ & $9.59\times10^{3}$ & $48.2$ &$2.87\times10^{-16}$   \\ 
		&$100 \times 100$  & $5.02 \times10^{2}$& $-$ & $-$ &$-$   \\ 
		 \midrule
		 \multirow{4}{*}{0.6}
		 &$40 \times 40$   & $7.20\times10^{0}$ & $1.39\times10^{2}$ & $19.3$  & $7.30\times10^{-16}$   \\ 
		&$60 \times 60$  & $5.91\times10^{1}$ & $1.66\times10^{3}$  & $28.1$ &$5.89\times10^{-16}$   \\ 
		&$80 \times 80$  & $1.99\times10^{2}$ & $9.79\times10^{3}$ & $49.2$ &$4.65\times10^{-16}$   \\ 
		&$100 \times 100$  & $5.07 \times10^{2}$& $-$ & $-$ &$-$   \\ 
		 \midrule
		 \multirow{4}{*}{0.8}
		 &$40 \times 40$   & $7.18\times10^{0}$ & $1.38\times10^{2}$ & $19.2$  & $1.01\times10^{-15}$   \\ 
		&$60 \times 60$  & $5.92\times10^{1}$ & $1.67\times10^{3}$  & $28.2$ &$8.09\times10^{-16}$   \\ 
		&$80 \times 80$  & $1.98\times10^{2}$ & $1.03\times10^{4}$ & $52.0$ &$6.88\times10^{-16}$   \\ 
		&$100 \times 100$  & $5.04 \times10^{2}$& $-$ & $-$ &$-$   \\ 
		\bottomrule
	\end{tabular}
	\caption{Horse images task with FGW metric. The comparison between FGC-FGW and the original algorithm with different $N$ and $\theta$. A dash means the running time exceeds 10 hours.}
	\label{tab:horse}
\end{table}

\section{Conclusion} \label{sec:conclusion}
In this paper, we demonstrate a special recursive relation over uniform grids and develop the Fast GW Gradient Computation (FGC). Thus, the computation of GW can be conducted in $O(N^2)$ time, which is almost optimal. 
Compared to the approximation or sampling-based methods, the fast algorithms produce full-sized and exact solutions as original entropic ones. Moreover, compared to other closed-form GW solutions over other special structures such as 1D real lines and trees, our method can be used to accelerate the computation of a wide scope of GW variants as long as the GW gradient is required, including unbalanced GW~\cite{sejourneUnbalancedGromovWasserstein2021}, Co-Optimal Transport~\cite{titouanCooptimalTransport2020}, and fixed support GW barycenter~\cite{peyreGromovWassersteinAveragingKernel2016}.
Empirical evaluations show that our FGC has a clear advantage in terms of time complexity and accelerates the cases from tens to hundreds of times. It is also validated that full-sized transport plans produced by our algorithm are exact under various settings. 
\section{Acknowledgements}
This work was supported by the National Natural Science Foundation of China (Grant No. $12271289$)$.$

\normalem
\bibliographystyle{unsrt}

\end{document}